# ON THE OPTIMAL DESIGN OF PARALLEL ROBOTS TAKING INTO ACCOUNT THEIR DEFORMATIONS AND NATURAL FREQUENCIES

**Sébastien Briot** [a, b]     **Anatol Pashkevich** [a, b, *]     **Damien Chablat** [a]

[a] Institut de Recherches en Communications et Cybernétique
de Nantes (IRCCyN), UMR CNRS 6597
1 rue de la Noë, BP 92101, 44321 Nantes Cedex 3 France

[b] Ecole des Mines de Nantes
La Chantrerie, 4 rue Alfred-Kastler, BP 20722,
44307 Nantes Cedex 3 France

**ABSTRACT**
This paper discusses the utility of using simple stiffness and vibrations models, based on the Jacobian matrix of a manipulator and only the rigidity of the actuators, whenever its geometry is optimised. In many works, these simplified models are used to propose optimal design of robots. However, the elasticity of the drive system is often negligible in comparison with the elasticity of the elements, especially in applications where high dynamic performances are needed. Therefore, the use of such a simplified model may lead to the creation of robots with long legs, which will be submitted to large bending and twisting deformations. This paper presents an example of manipulator for which it is preferable to use a complete stiffness or vibration model to obtain the most suitable design and shows that the use of simplified models can lead to mechanisms with poorer rigidity.

## 1 INTRODUCTION

Parallel robots have increasingly been used in industry since the last few years, mainly for pick-and-place applications or high-speed machining [1], [2]. This interest is due to their main properties, i.e. their higher rigidity and dynamic capacities compared with serial manipulators counterpart.

Clearly, having a good knowledge of the accuracy of a manipulator is a crucial point. The accuracy of a mechanism is due to several factors, as:
- manufacturing errors, which can however be taken into account through calibration;
- backlash, which can be eliminated through proper choice of mechanical components;
- active-joint errors, coming from the finite resolution of the encoders, sensor errors, and control errors, which may be reduced by using very accurate sensors
- rigidity of the mechanism, which may be improved through the use of more rigid structures. However, this would increase inertia, which is unacceptable in many applications, as in high-speed machining where operating speed is a crucial point, or for space operations where the embedded masses in the launcher should be minimized.

Thus, it is necessary at the first design stages to optimize the geometry, as well as the shape of the elements of the manipulator. This will lead to the creation of a mechanism, which will deform, or vibrate, but as few as possible. Therefore it is obvious that the designer should use stiffness models (elastostatic or elastodynamic) of the manipulator.

Several models have been proposed and used in the literature in order to compute the deformations and the natural frequencies of a mechanism. We may classify them into two principal groups:
- the simplified models based on the Jacobian matrix of the mechanism [3]-[8] which take only into consideration the elasticity of the actuators (i.e. the elasticity of the control loop – negligible if a PID control is implemented – plus the mechanical transmission system)
- more refined models taking into account the elasticity of all the links of the manipulator [9]- [16].

---

* Professor and author of correspondence, Phone: +33 2 51 85 83 00, Fax: +33 2 51 85 81 99, Email: Anatol.Pashkevich@emn.fr.

Most of papers dealing with the minimisation of the deformations of parallel robots for a desired task are using the simplified models based on the Jacobian matrix. The authors consider that the deformations (or vibrations) due to the elasticity in the actuators are preponderant with respect to the stiffness of a manipulator. However, we can notice that without considering the elasticity of its elements, the optimisation can lead to the creation of structures with longer legs, potentially submitted to bending or twisting, which will have much more large deformations than those due to the drive system. For example, in [5], the optimisation of the geometry of $\underline{P}R\underline{R}RP$[1] mechanisms is presented. It is shown that, for equivalent size of the workspace, the manipulators with the best stiffness are those that have the longer lengths of legs and larger bases. However, as the legs are submitted to flexure deformations along the axis orthogonal to the plane of the robot, longer the legs, more important the deformations. To only way to solve this problem is to increase the stiffness of the element, which reduces its dynamic capacities.

The same problem will appear when optimizing a manipulator for reducing its vibrations. For example, in [7], the authors propose to find the optimal geometry of a 2-degrees-of-freedom (DOF) manipulator dedicated for space applications. This manipulator is similar to a Gough-Stewart platform for which four of its actuators are replaced by rigid links. In this paper, the authors try to find the optimal geometry of the mechanism, which will maximize the natural frequencies by taking only into account the longitudinal vibrations of the legs of the robot. However, as pointed out in [17], the natural frequencies due to longitudinal vibrations of a beam are many times smaller, even negligible, than the natural frequencies due to the transverse (bending) vibrations (which are not taken into account in the considered work). Therefore, the poorest natural frequency of the system under study may have not been taken into account and, even so, may have been deteriorated. Therefore, as it will be shown in this paper, it is much more important to consider the elasticity into all the elements of the robot.

The purpose of this paper is to warn the scientific community about the necessity of having more complete models for computing the stiffness and the vibrations when designing a parallel manipulator. So, in this perspective, we will present the analyses of the elastostatic behaviour and of the natural frequencies of a prototype, designed at the IFW Institute in the context of the European NEXT project, using two models and comparing them, e.g. (i) simplified models based on the Jacobian matrix of the robot [4], [8] and (ii) more refined model based on the use of virtual joints [16], [18].

This paper will be organised as follows. In section 2, the elastostatic analysis of the IFW demonstrator is presented. In section 3, the natural frequencies of the mechanism are computed. In both sections, the comparison between the two models is presented. We also achieve a parametric analysis to study the influence of the geometry of the mechanism on its rigidity and to improve it by a proper design. Finally, conclusions are drawn in section 4.

---
[1] In the following of the paper, $P$ and $R$ ($\underline{P}$ and $\underline{R}$, resp.) stand for passive (active, resp.) prismatic and revolute pairs.

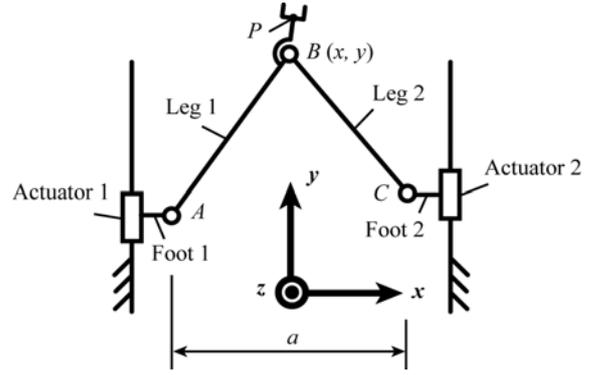

(a) Schematics.

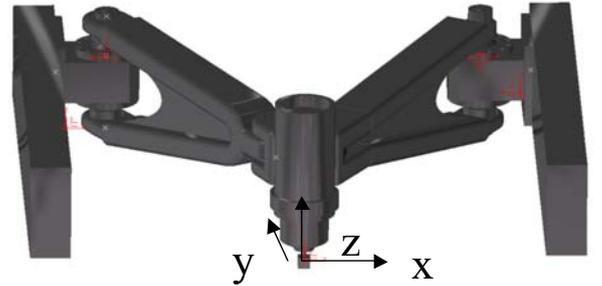

(b) CAD View.

**Figure 1.** Architecture of the $\underline{P}R\underline{R}RP$ robot under study.

## 2  STIFFNESS ANALYSIS

In this section, we will compare the deformations due to the elastostatic behaviour of the IFW demonstrator within its workspace. In the first part, we will take only into account the elasticity in the actuated joints (i.e. the elasticity of the drive transmission). In the second part, we consider the elasticity of all the robot links. In the third part, we will present a parametric analysis of the deformations of the robot.

### 2.1  Simplified Model

This model is the simplest model we may use. It states that the deformations $\delta\mathbf{t}$ of the end-effector are related to the efforts $\mathbf{f}$ applied on it and to the elasticity $K_i$ of the drive system $i$ via the well-known relation [3]:

$$\delta\mathbf{t} = \mathbf{J}\mathbf{K}^{-1}\mathbf{J}^T\mathbf{f} \qquad (1)$$

with $\mathbf{K} = diag(K_1, \ldots, K_n)$, $n$ being the number of actuators and $\mathbf{J}$ is the Jacobian matrix of the manipulator, relating the twist $\mathbf{t}$ of the platform to the actuators velocities $\dot{\mathbf{q}}$ as follows:

$$\mathbf{t} = \mathbf{J}\dot{\mathbf{q}}. \qquad (2)$$

The IFW demonstrator (Fig. 1) is a 3-axis milling machine which is composed of a planar parallel module (a $\underline{P}R\underline{R}RP$ robot, also called a Biglide) which allows the planar translational displacements of the tool, mounted in series with a linear vertical actuator which achieves the translations along the $z$ axis. This machine is designed for high-speed machining in aeronautic applications.

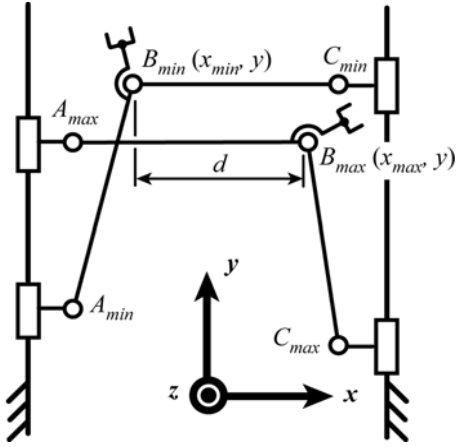

**Figure 2.** Maximal stroke along *x*-axis of the PRRRP robot.

Its planar module is a PRRRP robot of which axes of the prismatic pairs are parallel (Fig. 1a). It is composed of two feet linking linear actuators to the moving legs. The feet are rigidly attached to the actuators and linked to the legs by revolute joints. Moreover, the two legs are linked together by a revolute joint at $B$, of which coordinates along $x$ and $y$ axes are denoted as $x$ and $y$, respectively. In the following of this paper, let us denote as $a$ the distance along the $x$-axis between the centre of the revolute joints $A$ and $C$, and as $L_1$ (resp. $L_2$) the distance between the revolute joint centres $A$ and $B$ (resp. $B$ and $C$).

Due to the parallelism of the axes of the prismatic pairs, the position along $y$ axis is unlimited (as far as the stroke of the prismatic pair is unlimited). Moreover, the performances of the manipulator will not depend on its position along $y$ axis. Therefore, its properties will be examined as a function of parameter $x$ only. Its geometric, inertia and elastic parameters (computed using CAD and FEA softwares [16]) are listed in the appendix.

On this prototype, the designers have used linear actuators without transmission systems. Therefore, the drive system is considered to be infinitely rigid, but for our computation, we will make the assumption the matrix **K** of equation (2) will contain the terms corresponding to the translational stiffness of the feet along $y$ axis, which are about $1.10^9$ N/m. So now we will compute the deformations of the mechanism under a constant load. The workspace used for the computation is the stroke $d$ of point $B$ along the $x$-axis, which is bounded by the positions $B_{min}$ and $B_{max}$, which are Type 1 singular configurations of the manipulator [19] (Fig. 2).

Figure 3 represents the deformations of the manipulator computed with this simplified model. The curve in full line represents the norm of the planar deformations under a load $\mathbf{f_x}$ equal to $[1000\ N, 0, 0]^T$ and the one in dotted line the norm of the planar deformations under a load $\mathbf{f_y} = [0, 1000\ N, 0]^T$. The first observation is that these deformations are inferior to 1.1 µm. Moreover, on the boundaries of the workspace, the deformations due to the force $\mathbf{f_x}$ are equal to zero. This is due to the fact that, in such a configuration, because of the horizontal position of one of the leg, a force along the $x$-axis is completely transmitted to the base without involving any actuator efforts (Fig. 2). On the contrary, a force along $y$-axis will only be supported by one actuator. Therefore the deformations along $y$-axis are maximal. However, though the use of this model, the deformations along $z$ axis, such as the small rotations of the system, are not considered. It will be shown in the next part that they are preponderant contrary to the deformations into the plane $xOy$.

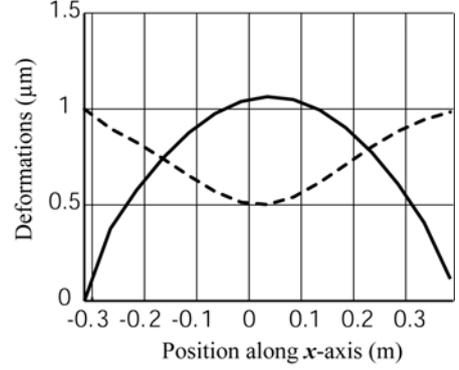

**Figure 3.** Planar deformations of the IFW demonstrator due to the forces $\mathbf{f_x}$ (full line) and $\mathbf{f_y}$ (dotted line) – simplified model.

## 2.2 Refined Lumped Model.

The modelisation used in this part has been presented by some of the authors in [16]. This model, which combines advantages of the traditional methods (the finite element analysis [9], [10], the matrix structural analysis [11], [12] and the virtual joint method [13], [14]) is based on a multidimensional lumped-parameter model that replaces the link flexibility by localized 6-DOF virtual springs that describe both the linear/rotational deflections and the coupling between them. The spring stiffness parameters are evaluated using FEA-modelling to ensure higher accuracy. In addition, it employs a new solution strategy of the kinetostatic equations, which allows computing the stiffness matrix for the overconstrained architectures, including the singular manipulator postures. This gives almost the same accuracy as FEA but with essentially lower computational effort because it eliminates the model re-meshing through the workspace.

This model states that the deformations $\delta\mathbf{t}_i$ of the extremity of the leg $i$ of the manipulator are related to the efforts $\mathbf{f}_i$ applied on its extremity via the relation:

$$\begin{bmatrix} \mathbf{S}_\theta^i & \mathbf{J}_q^i \\ \mathbf{J}_q^{iT} & \mathbf{0} \end{bmatrix} \begin{bmatrix} \mathbf{f}_i \\ \delta\mathbf{q}_i \end{bmatrix} = \begin{bmatrix} \delta\mathbf{t}_i \\ \mathbf{0} \end{bmatrix}, \quad \mathbf{S}_\theta^i = \mathbf{J}_\theta^i \left(\mathbf{K}_\theta^i\right)^{-1} \mathbf{J}_\theta^{iT} \quad (3)$$

where $\delta\mathbf{q}_i$ represents the passive joints displacements of the leg $i$, $\mathbf{K}_\theta^i$ is the stiffness matrix corresponding to the rigidity of all the elements of the leg $i$ and $\mathbf{J}_\theta^i$, $\mathbf{J}_q^i$ are the Jacobian matrices relating the displacements of the extremity of the leg $i$ to the spring deflections $\delta\mathbf{\theta}_i$ and passive joint displacements $\delta\mathbf{q}_i$, such as

$$\delta\mathbf{t}_i = \mathbf{J}_q^i \delta\mathbf{q}_i + \mathbf{J}_\theta^i \delta\mathbf{\theta}_i, \quad \mathbf{J}_q^i = \left[\frac{\partial \mathbf{t}_i}{\partial \mathbf{q}_i}\right], \quad \mathbf{J}_\theta^i = \left[\frac{\partial \mathbf{t}_i}{\partial \mathbf{\theta}_i}\right]. \quad (4)$$

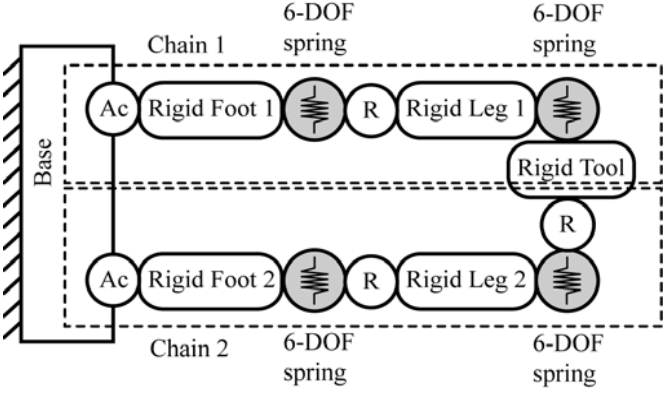

**Figure 4.** Flexible elastostatic model of the *PRRRP* robot (Ac: Actuated joint, R: *R* joint).

Matrices $\mathbf{K}_\theta^i$, $\mathbf{J}_q^i$ and $\mathbf{J}_\theta^i$ may be obtained through the following approach. Let us consider the flexible model of the robot at Fig. 4. The method presented in [16] states that each leg may be decomposed into a sequence of rigid links and virtual 6-DOF springs, which includes:

(a) a rigid link between the manipulator base and the *i*th actuating joint (part of the base platform) described by the constant homogenous transformation matrix $\mathbf{T}_{\text{Base}}^i$;

(b) a 1-DOF actuating joint which is defined by the homogenous matrix function $\mathbf{V}_a(q_0^i)$ where $q_0^i$ is the actuated coordinate;

(c) a rigid foot, which is described by the constant homogenous transformation $\mathbf{T}_{\text{Foot}}^i$;

(d) a 6-DOF virtual spring describing the foot stiffness, which is defined by the homogenous matrix function $\mathbf{V}_s(\theta_0^i,...,\theta_5^i)$ where $\{\theta_0^i,\theta_1^i,\theta_2^i\}$, $\{\theta_3^i,\theta_4^i,\theta_5^i\}$ are the virtual spring coordinates corresponding to the spring translational and rotational deflections;

(e) a 1-DOF passive *R*-joint at the beginning of the leg allowing one rotation with angle $q_1^i$, which is described by the homogenous matrix function $\mathbf{V}_{r1}(q_1^i)$;

(f) a rigid leg linking the foot and the end-effector, which is described by the constant homogenous transformation $\mathbf{T}_{\text{Leg}}^i$;

(g) a 6-DOF virtual spring describing the leg stiffness, which are defined by the homogenous matrix function $\mathbf{V}_s(\theta_6^i,...,\theta_{11}^i)$ where $\{\theta_6^i,\theta_7^i,\theta_8^i\}$ and $\{\theta_9^i,\theta_{10}^i,\theta_{11}^i\}$ are the spring translational/rotational deflections;

(h) a 1-DOF passive *R*-joint at the end of the leg 2 (not for the leg 1) allowing one rotation with angle $q_2^2$, which is described by the homogenous matrix function $\mathbf{V}_{r2}(q_2^2)$;

(i) a rigid link from the manipulator leg to the end-effector (part of the movable platform) described by the constant homogenous matrix transformation $\mathbf{T}_{\text{Tool}}^i$.

From these assumptions, the stiffness matrix of the robot under study has the following form [16]:

$$\mathbf{K}_\theta^i = \begin{bmatrix} \mathbf{K}_{\text{Foot}}^i & \mathbf{0}_{6\times 6} \\ \mathbf{0}_{6\times 6} & \mathbf{K}_{\text{Leg}}^i \end{bmatrix}. \tag{5}$$

For obtaining matrices $\mathbf{J}_q^i$ and $\mathbf{J}_\theta^i$, let us now consider the corresponding mathematical expression defining the end-effector location subject to variations of all above defined coordinates of a single kinematic chain *i* may be written as follows:

$$\mathbf{T}^1 = \mathbf{T}_{\text{Base}}^1 \mathbf{V}_a(q_0^1) \mathbf{T}_{\text{Foot}}^1 \mathbf{V}_s(\theta_0^1,...,\theta_5^1) \mathbf{V}_{r1}(q_1^1) \mathbf{T}_{\text{Leg}}^1 \\ \cdot \mathbf{V}_s(\theta_6^1,...,\theta_{11}^1) \mathbf{T}_{\text{Tool}}^1 \tag{6a}$$

$$\mathbf{T}^2 = \mathbf{T}_{\text{Base}}^2 \mathbf{V}_a(q_0^2) \mathbf{T}_{\text{Foot}}^2 \mathbf{V}_s(\theta_0^2,...,\theta_5^2) \mathbf{V}_{r1}(q_1^2) \mathbf{T}_{\text{Leg}}^2 \\ \cdot \mathbf{V}_s(\theta_6^2,...,\theta_{11}^2) \mathbf{V}_{r2}(q_2^2) \mathbf{T}_{\text{Tool}}^2 \tag{6b}$$

where the matrix function $\mathbf{V}_a(\cdot)$ is an elementary translation along *y*, the matrix functions $\mathbf{V}_{rj}(\cdot)$ (*j* = 1, 2) are elementary rotations around *z*, the spring matrix $\mathbf{V}_s(\cdot)$ is composed of six elementary transformations.

The matrix $\mathbf{J}_\theta^i$ may be obtained from the derivation of the matrix $\mathbf{T}^i$ with respect to the spring parameters $\theta_j^i$ (*j* = 0 to 11), at the point $\theta_j^i = 0$, considering that

$$\frac{\partial \mathbf{T}^i}{\partial \theta_j^i} = \mathbf{H}_{ij}^L \frac{\partial \mathbf{V}_{\theta_j}}{\partial \theta_j^i}(\theta_j^i)\mathbf{H}_{ij}^R = \begin{bmatrix} 0 & -\varphi'_{iz} & \varphi'_{iy} & p'_{ix} \\ \varphi'_{iz} & 0 & -\varphi'_{ix} & p'_{iy} \\ -\varphi'_{iy} & \varphi'_{ix} & 0 & p'_{iz} \\ 0 & 0 & 0 & 0 \end{bmatrix}. \tag{7}$$

where the first and the third multipliers are the constant homogenous matrices which do not include the displacement $\theta_j^i$, and the second multiplier corresponds to the derivative of the elementary translation or rotation corresponding to $\theta_j^i$. In the right-hand term, symbol " ' " stands for the derivation of the variables with respect to $\theta_j^i$. Therefore, $p'_{ix}$, $p'_{iy}$ and $p'_{iz}$ (resp. $\varphi'_{ix}$, $\varphi'_{iy}$ and $\varphi'_{iz}$) correspond to the small translations along (resp. rotations about) *x*, *y* and *z* axes of the extremity of the leg *i* due to the variation of the parameter $\theta_j^i$.

The Jacobians $\mathbf{J}_q^i$ can be computed in a similar manner, but the derivatives are evaluated in the neighborhood of the "nominal" values of the passive joint coordinates $q_j^i$ (*i* = 1, 2 if *i* = 2, *i* = 1 if not) corresponding to the rigid case (these values are obtained from the inverse kinematics.

Finally, the stiffness matrix $\mathbf{K}_i$ of the leg *i*, which relates the deformations $\delta\mathbf{t}_i$ to the force $\mathbf{f}_i$ as

$$\mathbf{f}_i = \mathbf{K}_i\,\delta\mathbf{t}_i, \tag{8}$$

can be computed by direct inversion of relevant 7 by 7 (8 by 8 in the case of the leg 2) matrix in the left hand side of (3) and extracting the 6 by 6 sub-matrix with indices corresponding to $\mathbf{S}_\theta^i$.

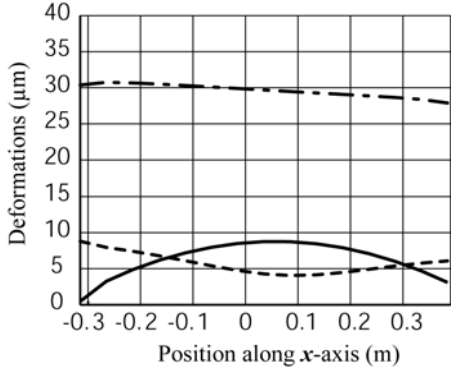

**Figure 5.** Deformations of the IFW demonstrator due to the forces $\mathbf{f_x}$ (full line), $\mathbf{f_y}$ (dotted line) and $\mathbf{f_z}$ (dashed line) – refined lumped model.

After the stiffness matrices $\mathbf{K}_i$ for all kinematic chains are computed, the stiffness of the entire manipulator can be found by simple addition:

$$\mathbf{K}_m = \sum_{i=1}^{n} \mathbf{K}_i . \qquad (9)$$

So now we will compute the deformations of the mechanism within its workspace under a constant load (Fig. 5). On this picture, the curves in full line represent the norm of the planar deformations due to the force $\mathbf{f_x}$ (applied at the extremity of the robot), the curves in dotted line the planar deformations due to a force $\mathbf{f_y}$, and the curve in dashed line the deformations along $z$ axis due to a force $\mathbf{f_z}$ equal to $[0, 0, 1000\ N]^T$. We may see that the deformations along $z$ axis are 3 times larger than those along $x$ and $y$ axes. Moreover, the planar deformations are about 10 times larger than with the simplified model. This is normal because the simplified model does not take into account the deformations of the legs. However, the curves keep the same profile. Therefore, the simplified model of section 2.1 gives a good idea of how will be the planar deformations of the mechanism inside its workspace.

In the next part, it will be shown that optimizing the geometry of the mechanism when considering only the simplified model will lead to large and preponderant deformations along $z$ axis.

### 2.3 Parametric Analysis.

In this part, we would like to analyze the deformations of the structure in several points of the workspace when the geometric parameters of the mechanism are changing. However, due to the complexity of the shape of the legs, it is quite complicated to express their stiffness matrix as a function of their lengths.

It is well known that the stiffness matrix of a beam may be written using symbolic expressions [16]. So, in order to make a parametric analysis, we replace the real legs of the robot by beams of constant cross sections that have the same four first diagonal terms as the stiffness matrices of the initial elements. It is obvious that the elements of the final mechanism will not be designed like beams. More efficient shapes could be drawn. But the aim of this part is to show that, for certain applications, (such as high-speed machining for which the mass of the structure is a crucial factor which decreases the

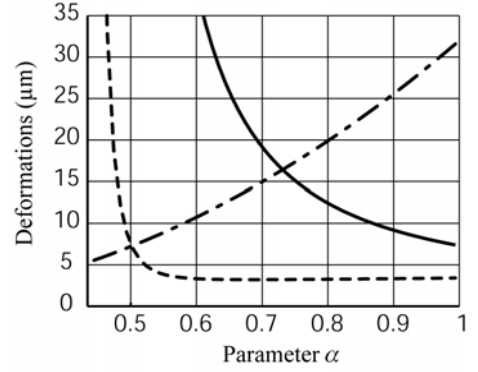

(a) at $x = a - L_2 + d/2$ (centre of the workspace).

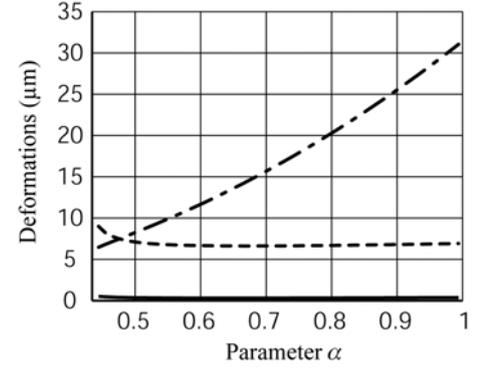

(b) at $x = a - L_2$ (left extremity of the workspace).

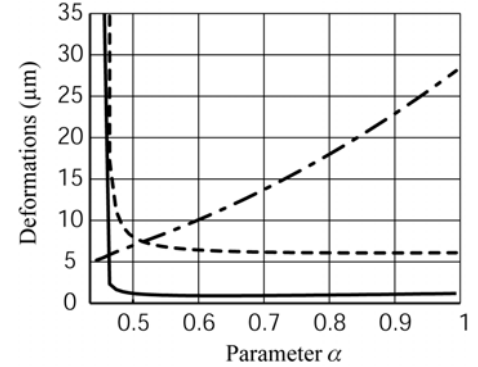

(c) at $x = L_1 - a$ (right extremity of the workspace).

**Figure 6.** Deformations of the IFW demonstrator due to the forces $\mathbf{f_x}$ (full line), $\mathbf{f_y}$ (dotted line) and $\mathbf{f_z}$ (dashed line) as functions of the parameter $\alpha$.

acceleration capacities and for which the stiffness of the elements may not be infinitely increased) there will always be a length of the legs for which, even is the mechanism is foreseen to be the stiffest with the simplified model, indeed, its stiffness will be poorer along the $z$ axis. And for such a preliminary analysis, the proposed parametric approach is completely sufficient.

So, the next step is to vary the lengths of the elements and to compute the deformations under the application of (i) planar efforts and (ii) a vertical force. But, in order to make a fair comparison, all manipulators should have the same workspace, i.e. the length of the elements should ensure that the maximal course $d$ ($d = L_1 + L_2 - a$, Fig. 2) along $x$ axis should be constant. However, *d* depends on three different parameters. However, to simplify the preliminary analysis, we consider that the new lengths of the legs will be a multiple of the lengths of the initial mechanism, denoted as $L_{10}$ and $L_{20}$. So, we define a dimensionless parameter $\alpha$ of which expression will be:

$$\alpha = \frac{L_1}{L_{10}} = \frac{L_2}{L_{20}} . \quad (10)$$

So, the parameter *a* may be defined as
$$a = \alpha (L_{10} + L_{20}) - d . \quad (11)$$

Finally, we plot the deformations of the robot (Fig. 6). On these pictures, the curves in full line represent the norm of the planar deformations due to the force $\mathbf{f_x}$, the curve in dotted line the planar deformations due to the force $\mathbf{f_y}$, and in dashed line the deformations along *z* axis due to the force $\mathbf{f_z}$ (the deformations along the other directions are null). These pictures show that, the longer the legs, the smaller the planar deformations, as presented in [5]. However, in such cases, the deformations along *z* axis will be increased. Therefore, the lengths of the legs that will minimize the global deformations of the mechanism will be smaller than that foreseen in the design optimisation using the simplified stiffness model.

## 3  NATURAL FREQUENCIES ANALYSIS

We will now compare the natural frequencies of the IFW demonstrator within its workspace. Natural frequencies will indicate the way a mechanism tends to vibrate. Moreover, the first natural frequency is associated with the highest level of energy due to vibrations, and represents the highest displacements of the structure. Therefore the lowest natural frequency is a good indicator of the dynamic performances of a mechanism. Please note that, in general, it is considered that the first natural frequency has to be out of the range of normal use of the machine, and especially for machine tools, greater than 100 Hz.

In the first section, we will compute the natural frequencies of the IFW demonstrator, taking only into account the elasticity in the actuated joints and the mass of the platform. In the second section, we present a more complete model that takes into consideration the elasticity and the inertia parameters of all the robot links. In the third section, we will present a parametric analysis.

### 3.1  Simplified Model

This model has been presented in [8]. It states that the natural frequencies $f_i$ of the system, taking only into account the elasticity of the drive system and the mass and inertia of the platform, are the solutions of the equation:

$$\det(\mathbf{J}^{-1} \mathbf{K} \mathbf{J}^{-T} - \omega_i^2 \mathbf{M}) = 0 , \quad \omega_i = 2\pi f_i \quad (12)$$

with **M** the inertia matrix of the platform [8].

The stiffness of the drive system is considered, as previously, to be equal to $1.10^9$ N/m. The mass of the tool is of 46 kg. So now we will compute the first natural frequency of the mechanism within its workspace (Fig. 7). It is shown that the first natural frequency is superior to 710 Hz, which is very high when considering that the first natural frequency should be superior to 100 Hz. However, though the use of this model, the vibrations along *z* axis are not considered. It will be presented in the next part that they are preponderant contrary to the vibrations into the plane *xOy*.

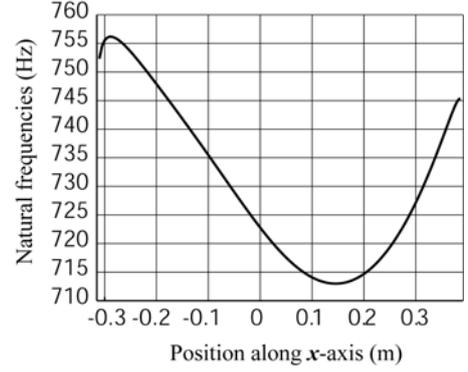

**Figure 7.** First natural frequency of the IFW demonstrator – simplified model.

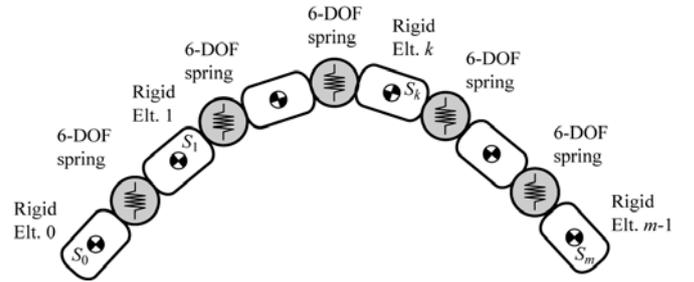

**Figure 8.** Division of a system into *m* rigid elements and *m* -1 spring elements.

### 3.2  Refined Lumped Model

The modelisation used in this part has been presented in [18]. This model, based on the theory of beams, states that each link of the mechanism may be replaced by a discrete number of rigid elements linked together by virtual springs. In [18], however, the authors consider that each spring may be represented by a diagonal stiffness matrix. But, this does not take into consideration the coupling between the translations and the rotations for the repartition of the efforts into the beam. Therefore, we propose to improve this model by considering, as for the elastostatic modelisation, springs represented by 6-DOF non-diagonal stiffness matrices.

Let us consider that any link may be decomposed into *m* rigid elements linked together by *m* – 1 6-DOF springs (Fig. 8). Its kinetic energy, denoted as *T*, may be expressed as:

$$2T = \sum_{k=0}^{m-1} \dot{\mathbf{q}}_k^T \mathbf{M}_k \dot{\mathbf{q}}_k \quad (13)$$

where $\dot{\mathbf{q}}_k = [\dot{x}_k, \dot{y}_k, \dot{z}_k, \dot{\varphi}_k^x, \dot{\varphi}_k^y, \dot{\varphi}_k^z]^T$, where $[\dot{x}_k, \dot{y}_k, \dot{z}_k]^T$ and $[\dot{\varphi}_k^x, \dot{\varphi}_k^y, \dot{\varphi}_k^z]^T$ are the translational and rotational velocities of the centre of masses $S_k$ of the element *k*, and

$$\mathbf{M}_k = \mathbf{D}_k \begin{bmatrix} m_k \mathbf{I}_3 & \mathbf{0}_{3\times 3} \\ \mathbf{0}_{3\times 3} & \mathbf{J}_k \end{bmatrix} \mathbf{D}_k^T , \quad \mathbf{D}_k = \begin{bmatrix} \mathbf{R}_k & \mathbf{0}_{3\times 3} \\ \mathbf{0}_{3\times 3} & \mathbf{R}_k \end{bmatrix} \quad (14)$$

with $\mathbf{I}_3$ and $\mathbf{0}_{3\times 3}$ the 3 by 3 identity and zero matrices, $m_k$ and $\mathbf{J}_k$ the mass and the inertia matrix of element *k*. $\mathbf{R}_k$ is the transformation matrix representing the rigid rotation of the element *k* with respect to the base frame.

The potential energy of the link, denoted as *V*, may be expressed as:

$$2V = \sum_{k=0}^{m-2} \boldsymbol{\theta}_k^T \mathbf{K}_k \boldsymbol{\theta}_k \,, \quad \mathbf{K}_k = \mathbf{D}_k \mathbf{K}_k^s \mathbf{D}_k^T \qquad (15)$$

where $\boldsymbol{\theta}_k = [\theta_0^k, \theta_1^k, \theta_2^k, \theta_3^k, \theta_4^k, \theta_5^k]^T$, $\{\theta_0^k, \theta_1^k, \theta_2^k\}$, $\{\theta_3^k, \theta_4^k, \theta_5^k\}$ are the virtual spring coordinates corresponding to the spring translational and rotational deflections of the element *k*, and $\mathbf{K}_k^s$ is the stiffness matrix of the spring *k*.

As the elements composing the beam are rigid, the relation linking the spring deflections $\boldsymbol{\theta}_k$ to the displacements of the centres of masses of the rigid elements $\mathbf{q}_k$ and $\mathbf{q}_{k+1}$ is:

$$\boldsymbol{\theta}_k = \begin{bmatrix} \mathbf{C}_{2,k} & \mathbf{C}_{1,k+1} \end{bmatrix} \begin{bmatrix} \mathbf{q}_k \\ \mathbf{q}_{k+1} \end{bmatrix} \qquad (16)$$

$\mathbf{C}_k$ and $\mathbf{C}_{k+1}$ being 6 by 6 matrices of which expressions are:

$$\mathbf{C}_{2,k} = -\begin{bmatrix} 1 & 0 & 0 & 0 & 0 & 0 \\ 0 & 1 & 0 & 0 & 0 & d_{2,k} \\ 0 & 0 & 1 & 0 & -d_{2,k} & 0 \\ 0 & 0 & 0 & 1 & 0 & 0 \\ 0 & 0 & 0 & 0 & 1 & 0 \\ 0 & 0 & 0 & 0 & 0 & 1 \end{bmatrix} \qquad (17)$$

$$\mathbf{C}_{1,k+1} = \begin{bmatrix} 1 & 0 & 0 & 0 & 0 & 0 \\ 0 & 1 & 0 & 0 & 0 & -d_{1,k+1} \\ 0 & 0 & 1 & 0 & d_{1,k+1} & 0 \\ 0 & 0 & 0 & 1 & 0 & 0 \\ 0 & 0 & 0 & 0 & 1 & 0 \\ 0 & 0 & 0 & 0 & 0 & 1 \end{bmatrix} \qquad (18)$$

where $d_{2,k}$ and $d_{1,k+1}$ are the distances between the centre of the spring and the centres of masses of elements *k* and *k*+1, respectively. Please note that this expression may be obtained from the well-known relation, assuming that all rotations are small (i.e. $\sin t = t$ and $\cos t = 1$)

$$\boldsymbol{\theta}_k = \begin{bmatrix} \mathbf{I}_3 & \mathbf{0}_{3\times 3} \\ \mathbf{0}_{3\times 3} & \mathbf{R}_{k+1}^s \end{bmatrix} \mathbf{q}_{k+1} - \begin{bmatrix} \mathbf{I}_3 & \mathbf{0}_{3\times 3} \\ \mathbf{0}_{3\times 3} & \mathbf{R}_k^s \end{bmatrix} \mathbf{q}_k \qquad (19)$$

where $\mathbf{R}_k^s$ and $\mathbf{R}_{k+1}^s$ are matrices of small rotations,

$$\mathbf{R}_k^s = \begin{bmatrix} 0 & -\varphi_k^z & \varphi_k^y \\ \varphi_k^z & 0 & -\varphi_k^x \\ -\varphi_k^y & \varphi_k^x & 0 \end{bmatrix}, \text{ and} \qquad (20a)$$

$$\mathbf{R}_{k+1}^s = \begin{bmatrix} 0 & -\varphi_{k+1}^z & \varphi_{k+1}^y \\ \varphi_{k+1}^z & 0 & -\varphi_{k+1}^x \\ -\varphi_{k+1}^y & \varphi_{k+1}^x & 0 \end{bmatrix}. \qquad (20b)$$

Thus, relation (15) may be rewritten under the form

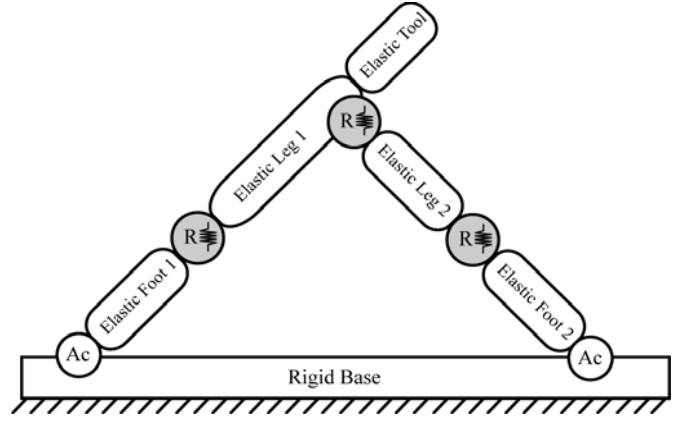

**Figure 9.** Flexible elastodynamic model of the _PRRRP_ robot (Ac: Actuated joint, R: elastic *R* joint).

$$2V = \sum_{k=1}^{m-2} \begin{bmatrix} \mathbf{q}_k^T, \mathbf{q}_{k+1}^T \end{bmatrix} \begin{bmatrix} \mathbf{C}_k^T \\ \mathbf{C}_{k+1}^T \end{bmatrix} \mathbf{K}_k \begin{bmatrix} \mathbf{C}_k & \mathbf{C}_{k+1} \end{bmatrix} \begin{bmatrix} \mathbf{q}_k \\ \mathbf{q}_{k+1} \end{bmatrix}. \qquad (21)$$

Differentiating the Lagrangian *L* of the link ($L = T - V$) to obtain the Lagrange equations leads to:

$$\frac{d}{dt}\left(\frac{\partial L}{\partial \dot{\mathbf{q}}}\right) - \frac{\partial L}{\partial \mathbf{q}} = \frac{d}{dt}\left(\frac{\partial T}{\partial \dot{\mathbf{q}}}\right) - \frac{\partial T}{\partial \mathbf{q}} + \frac{\partial V}{\partial \mathbf{q}} = \mathbf{M}\ddot{\mathbf{q}} + \mathbf{K}\mathbf{q} = \mathbf{0} \qquad (22)$$

where $\mathbf{q} = [\mathbf{q}_0, \mathbf{q}_1, \ldots, \mathbf{q}_k, \ldots \mathbf{q}_{m-1}]^T$, $\mathbf{M} = diag(\mathbf{M}_0, \ldots, \mathbf{M}_{m-1})$, $\mathbf{K} = \mathbf{C}^T diag(\mathbf{K}_0, \ldots, \mathbf{K}_{m-2})\mathbf{C}$, and

$$\mathbf{C} = \begin{bmatrix} \mathbf{C}_{2,0} & \mathbf{C}_{1,1} & \mathbf{0}_{6\times 6} & \cdots & \mathbf{0}_{6\times 6} & \mathbf{0}_{6\times 6} \\ \mathbf{0}_{6\times 6} & \mathbf{C}_{2,1} & \mathbf{C}_{1,3} & \cdots & \mathbf{0}_{6\times 6} & \mathbf{0}_{6\times 6} \\ \vdots & \vdots & \vdots & \ddots & \vdots & \vdots \\ \mathbf{0}_{6\times 6} & \mathbf{0}_{6\times 6} & \mathbf{0}_{6\times 6} & \cdots & \mathbf{C}_{2,m-2} & \mathbf{C}_{1,m-1} \end{bmatrix}. \qquad (23)$$

Finally, the natural frequencies $f_i$ of the link are the solutions of the equation:

$$\det(\mathbf{K} - \omega_i^2 \mathbf{M}) = 0, \quad \omega_i = 2\pi f_i. \qquad (24)$$

If the link is fixed at one of its extremity (here, for example, consider that the element 0 is rigidly linked to the base, i.e. $\mathbf{q}_0 = \mathbf{0}$), so the lines and raws with indices corresponding to $\mathbf{q}_0$ should be removed from matrices **M** and **K** before computing the natural frequencies.

Let us now assume that, for a given position of the tool, the robot may be considered as an assembly of several elastic links (modelised as on Fig. 8) of which extremity are linked to the base (Fig. 9). The connexions between these links (the passive *R* joints) may be considered as modified springs, of which stiffness around *z* axis is null, i.e. their stiffness matrices, denoted as $\mathbf{K}_R$, could be written under the form,

$$\mathbf{K}_R = \begin{bmatrix} \mathbf{I}_5 & \mathbf{0}_{5\times 1} \\ \mathbf{0}_{1\times 5} & 0 \end{bmatrix} \mathbf{K}_S \begin{bmatrix} \mathbf{I}_5 & \mathbf{0}_{5\times 1} \\ \mathbf{0}_{1\times 5} & 0 \end{bmatrix}, \qquad (25)$$

where $\mathbf{I}_5$ is the identity matrix of dimension 5, $\mathbf{0}_{ij}$ is a zero matrix with *i* lines and *j* raws, and $\mathbf{K}_S$ represents the stiffness matrix of the element considered linked to the *R* joint. So, with such assumptions, the same approach may be applied to compute the natural frequencies of the <u>PRRRP</u> robot under study.

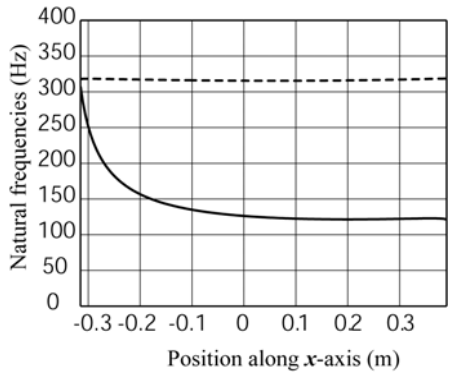

**Figure 10.** Natural frequencies of the IFW demonstrator – refined lumped model.

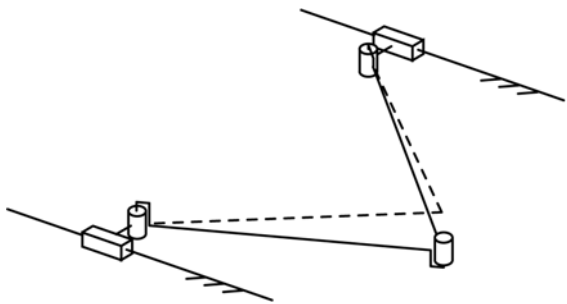

(a) at its first natural mode.

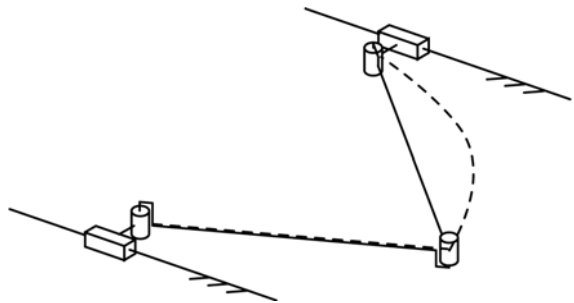

(b) at its second natural mode.

**Figure 11.** Schematics of the deformations of the robot.

Let us now compute the natural frequencies of the robot. For the computation, we have discretized the beams into 20 elements. On Fig. 10 are represented the two first natural frequencies of the mechanism. The full line is for the first mode and the dotted line for the second mode. The first mode corresponds to the bending of the robot along the $z$ axis (Fig. 11a). The second mode is a planar vibration, due to the bending vibration of the leg 1 (Fig. 11b). Moreover, analyzing the 60 first modes using a FEA software, no one is only due to the elasticity of the actuators.

Finally, in order to roughly correlate the results of Fig. 10, please note that the first natural frequency of an equivalent beam corresponding to leg 1 is about 180 Hz. The value found

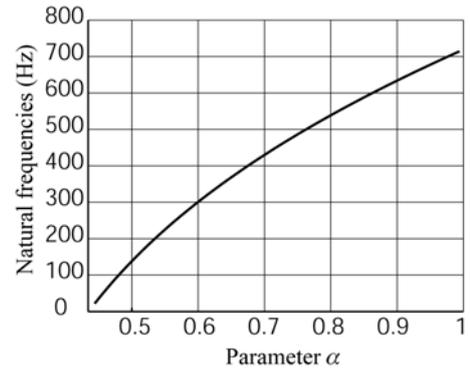

(a) at $x = a - L_2 + d/2$ (centre of the workspace).

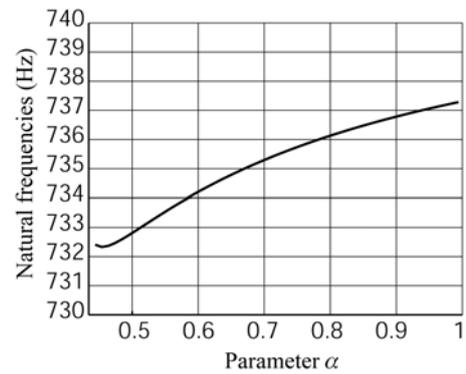

(b) at $x = a - L_2$ (left extremity of the workspace).

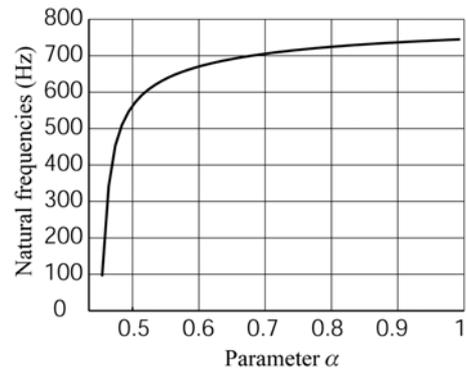

(c) at $x = L_1 - a$ (right extremity of the workspace).

**Figure 12.** First natural frequency of the IFW demonstrator as a function of the parameter $\alpha$ – simplified model.

with the lumped model, by modelizing the beam with 20 elements, is inferior to 1% to the theoretical result. These values are of the same order of those found on Fig. 10. This is quite realistic because, as the robot is made of an assembly of beams, the obtained global natural frequencies should not be very different from those of the beams of which it is composed.

In the next part, we will make a parametric analysis of the natural frequencies of the mechanism as a function of its geometric parameters.

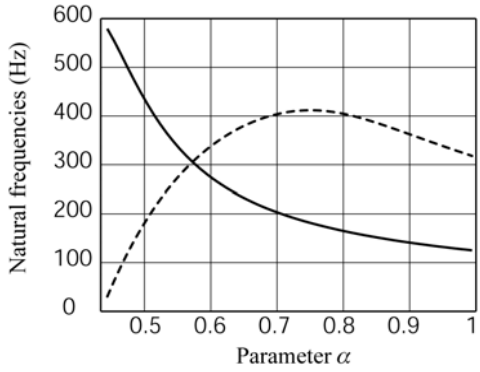

(a) at $x = a - L_2 + d/2$ (centre of the workspace).

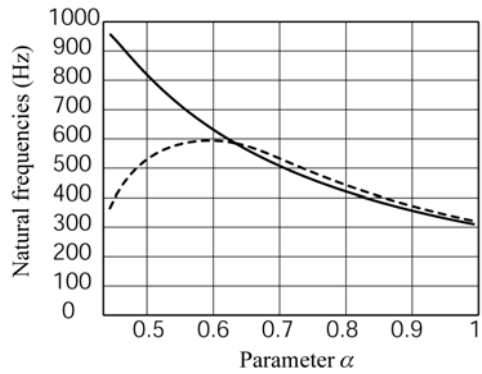

(b) at $x = a - L_2$ (left extremity of the workspace).

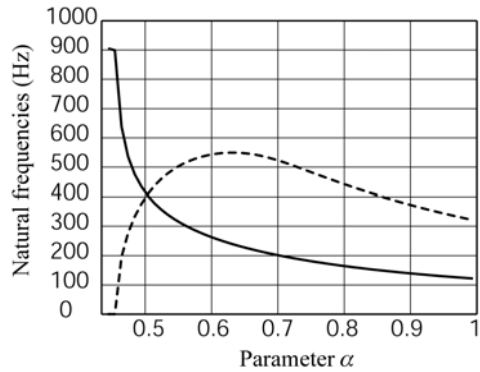

(c) at $x = L_1 - a$ (right extremity of the workspace).

**Figure 13.** First (full line) and second (dotted line) natural frequencies of the IFW demonstrator as a function of the parameter $\alpha$ – refined lumped model.

### 3.3 Parametric Analysis.

In this part, we would like to analyze the natural modes of the structure in several points of the workspace when the geometric parameters of the mechanism are changing. As previously, due to the complexity of the shape of the legs, it is quite complicated to express their stiffness and mass matrices as a function of their lengths.

So, in order to make a parametric analysis, we replace another time the real legs of the robot by beams of constant cross sections that globally have the same diagonal terms as the stiffness matrices of the initial elements.

As in the section 2.3, we vary the parameter $\alpha$ of eq. (6) and compute the natural frequencies of the mechanisms in their workspace. We also consider that the robots should have the same workspace, i.e. the value of the distance $a$ is given section 2.3. In a first step, we consider only the simplified model of section 3.1. On Fig. 12 is represented the first natural frequency of the robot within its workspace, as a function of the parameter $\alpha$. One may see that, longer the legs, higher the natural frequency, i.e. better the mechanism. These results are similar to those obtained with the simplified stiffness model presented in [5].

Then, we plot the natural frequencies of the robot using the model of section 3.2 (Fig. 13). On these pictures, the curves in full line represent the natural frequencies due to the bending of the entire robot (Fig. 11a), the curves in dotted line the natural frequencies due to the bending of leg 1 (Fig. 11b). We do not represent the other natural frequencies, as for this manipulator, they never are inferior to these two ones. These pictures show that, longer the legs, smaller the natural frequencies, which is in contradiction with the simplified model based on the Jacobian matrix. Indeed, the lengths of the legs that will minimize the global deformations of the mechanism will be smaller than that foreseen in the design optimisation using the simplified model.

### 4 CONCLUSIONS

The aim of this paper is to warn the scientific community about the utility of using simple stiffness and vibrations models, based on the Jacobian matrix of a manipulator and only taking into account the rigidity of the actuators, when optimizing its geometry. It has been shown that, in many works, these simplified models are used to propose optimal design of robots. However, the elasticity of the drive system is negligible in comparison with the elasticity of the elements, especially in applications where high dynamic performances are needed. Therefore, the use of such a simplified model may lead to the creation of robots with long legs, which will be submitted to large bending deformations.

This paper presented an example of *PRRRP* robot for which it is preferable to use a more complex stiffness or vibration model to obtain the most adequate design. It has been shown that the use of simplified models will lead to mechanisms with longer legs, for which bending deformations are preponderant, which will lead to a poorer rigidity of the robot. For this reason, it is obvious that more complex models, such as those presented in this paper, should be used in preliminary analyses, in order to define the best architecture, which will minimize the deformations of the end-effectors, as well as its vibrations.

### 5 ACKNOWLEDGMENT

This work has been partially funded by the French région Pays de la Loire and by the European project NEXT, acronyms for "Next Generation of Productions Systems", Project No. IP 011815.

## 7  APPENDIX

For the IFW manipulator, the lengths of the elements are given by: $a = 0.92$ m, $L_1 = 0.85$ m, $L_2 = 0.775$ m, $L_{Tool} = 0.155$.

The masses of the elements are: $m_{Leg1} = 69.705$ kg, $m_{Leg2} = 49.366$ kg and $m_{Tool} = 46$ kg. The mass of the tool is applied at point $P$.

We consider that the inertia of the tool is negligible compared with the inertia of the other elements. The inertia matrices of the links, expressed in the local frames attached to the links (the local $x$ axes are considered along the direction of the beams, the local $z$ axes are parallel to the global $z$ axis), at the centre of masses, are:

$$\mathbf{J}_{Foot} = \begin{bmatrix} 0.268 & 0 & 0 \\ 0 & 0.211 & 0 \\ 0 & 0 & 0.261 \end{bmatrix} \text{kg.m}^2 \quad \mathbf{J}_{Leg1} = \begin{bmatrix} 1.187 & -0.164 & -1.247 \\ -0.164 & 3.022 & -0.940 \\ -1.247 & -0.940 & 2.646 \end{bmatrix} \text{kg.m}^2$$

$$\mathbf{J}_{Leg2} = \begin{bmatrix} 6.122 & 0.014 & 0.312 \\ 0.014 & 5.848 & -0.314 \\ 0.312 & -0.314 & 0.635 \end{bmatrix} \text{kg.m}^2.$$

The position of the centre of masses of the legs along the local $x$ axes of the beams (from the centre of the revolute joints $A$ and $C$, respectively) is: $L_{G1} = 0.542$ m, $L_{G2} = 0.375$ m, where $G_i$ is the centre of masses of the leg $i$.

The link compliance matrices were computed via the FEA-based simulation technique presented in [16], which yielded

$$\mathbf{k}_{Foot} = \begin{bmatrix} 1.67\cdot10^{-10} & 8.85\cdot10^{-13} & -7.78\cdot10^{-14} & -2.12\cdot10^{-13} & 7.95\cdot10^{-12} & 2.50\cdot10^{-12} \\ 8.85\cdot10^{-13} & 5.87\cdot10^{-9} & 6.39\cdot10^{-12} & -3.58\cdot10^{-11} & -2.12\cdot10^{-11} & 3.94\cdot10^{-8} \\ -7.78\cdot10^{-14} & 6.39\cdot10^{-12} & 5.53\cdot10^{-10} & 1.35\cdot10^{-11} & -4.49\cdot10^{-9} & 1.91\cdot10^{-11} \\ -2.12\cdot10^{-13} & -3.58\cdot10^{-11} & 1.35\cdot10^{-11} & 6.96\cdot10^{-8} & 5.28\cdot10^{-11} & -2.71\cdot10^{-10} \\ 7.95\cdot10^{-12} & -2.12\cdot10^{-11} & -4.49\cdot10^{-9} & 5.28\cdot10^{-11} & 8.48\cdot10^{-8} & 7.40\cdot10^{-9} \\ 2.50\cdot10^{-12} & 3.94\cdot10^{-8} & 1.91\cdot10^{-11} & -2.71\cdot10^{-10} & 7.40\cdot10^{-9} & 3.16\cdot10^{-9} \end{bmatrix}$$

$$\mathbf{k}_{Leg1} = \begin{bmatrix} 2.81\cdot10^{-9} & -1.01\cdot10^{-8} & -1.41\cdot10^{-9} & -1.81\cdot10^{-9} & 4.42\cdot10^{-9} & 2.92\cdot10^{-8} \\ -1.01\cdot10^{-8} & 1.77\cdot10^{-7} & -1.83\cdot10^{-9} & -2.03\cdot10^{-9} & 2.93\cdot10^{-9} & -2.90\cdot10^{-7} \\ -1.41\cdot10^{-9} & -1.83\cdot10^{-9} & 3.19\cdot10^{-8} & 4.77\cdot10^{-8} & -9.94\cdot10^{-8} & -2.27\cdot10^{-9} \\ -1.81\cdot10^{-9} & -2.03\cdot10^{-9} & 4.77\cdot10^{-8} & 1.73\cdot10^{-7} & -8.02\cdot10^{-8} & -1.18\cdot10^{-9} \\ 4.42\cdot10^{-9} & 2.93\cdot10^{-9} & -9.94\cdot10^{-8} & -8.02\cdot10^{-8} & 8.13\cdot10^{-7} & 3.23\cdot10^{-8} \\ 2.92\cdot10^{-8} & -2.90\cdot10^{-7} & -2.27\cdot10^{-9} & -1.18\cdot10^{-9} & 3.23\cdot10^{-8} & 6.08\cdot10^{-7} \end{bmatrix}$$

$$\mathbf{k}_{Leg2} = \begin{bmatrix} 2.71\cdot10^{-10} & 1.29\cdot10^{-10} & -1.99\cdot10^{-10} & 4.68\cdot10^{-9} & 1.73\cdot10^{-9} & -7.06\cdot10^{-11} \\ 1.29\cdot10^{-10} & 1.26\cdot10^{-8} & -3.88\cdot10^{-13} & 1.84\cdot10^{-10} & 1.67\cdot10^{-8} & -2.12\cdot10^{-8} \\ -1.99\cdot10^{-10} & -3.88\cdot10^{-13} & 1.07\cdot10^{-9} & -1.03\cdot10^{-8} & -2.38\cdot10^{-10} & 3.71\cdot10^{-13} \\ 4.68\cdot10^{-9} & 1.84\cdot10^{-10} & -1.03\cdot10^{-8} & 2.52\cdot10^{-7} & 3.62\cdot10^{-9} & 4.54\cdot10^{-10} \\ 1.73\cdot10^{-9} & 1.67\cdot10^{-8} & -2.38\cdot10^{-10} & 3.62\cdot10^{-9} & 7.22\cdot10^{-7} & 3.95\cdot10^{-8} \\ -7.06\cdot10^{-11} & -2.12\cdot10^{-8} & 3.71\cdot10^{-13} & 4.54\cdot10^{-10} & 3.95\cdot10^{-8} & 1.70\cdot10^{-7} \end{bmatrix}$$

$$\mathbf{k}_{Tool} = \begin{bmatrix} 1.16\cdot10^{-9} & -9.70\cdot10^{-11} & -1.33\cdot10^{-11} & 6.88\cdot10^{-9} & 4.89\cdot10^{-8} & -2.64\cdot10^{-9} \\ 9.70\cdot10^{-11} & 1.33\cdot10^{-9} & -1.15\cdot10^{-10} & -5.91\cdot10^{-8} & -7.11\cdot10^{-9} & 1.96\cdot10^{-11} \\ -1.33\cdot10^{-11} & -1.15\cdot10^{-10} & 5.53\cdot10^{-10} & 2.30\cdot10^{-9} & 6.52\cdot10^{-10} & 2.00\cdot10^{-10} \\ 6.88\cdot10^{-9} & -5.91\cdot10^{-8} & 2.30\cdot10^{-9} & 3.77\cdot10^{-6} & 4.23\cdot10^{-7} & -2.87\cdot10^{-8} \\ 4.89\cdot10^{-8} & -7.11\cdot10^{-9} & 6.52\cdot10^{-10} & 4.23\cdot10^{-7} & 3.15\cdot10^{-6} & -6.94\cdot10^{-8} \\ -2.64\cdot10^{-9} & 1.96\cdot10^{-11} & 2.00\cdot10^{-10} & -2.87\cdot10^{-8} & -6.94\cdot10^{-8} & 3.29\cdot10^{-6} \end{bmatrix}$$